# Interpreting and Understanding Graph Convolutional Neural Network using Gradient-based Attribution Method

**SHANGSHENG XIE[1], MINGMING LU[1,*]**

[1] School of Computer Science and Engineering, Central South University, Changsha, 410083, China

Corresponding author: Mingming Lu (e-mail: mingminglu@csu.edu.cn).

**ABSTRACT** To solve the problem that convolutional neural networks (CNNs) are difficult to process non-grid type relational data like graphs, Kipf et al. proposed a graph convolutional neural network (GCN). The core idea of the GCN is to perform two-fold informational fusion for each node in a given graph during each iteration: the fusion of graph structure information and the fusion of node feature dimensions. Because of the characteristic of the combinatorial generalizations, GCN has been widely used in the fields of scene semantic relationship analysis, natural language processing and few-shot learning etc. However, due to its two-fold informational fusion involves mathematical irreversible calculations, it is hard to explain the decision reason for the prediction of the each node classification. Unfortunately, most of the existing attribution analysis methods concentrate on the models like CNNs, which are utilized to process grid-like data. It is difficult to apply those analysis methods to the GCN directly. It is because compared with the independence among CNNs input data, there is correlation between the GCN input data. This resulting in the existing attribution analysis methods can only obtain the partial model contribution from the central node features to the final decision of the GCN, but ignores the other model contribution from central node features and its neighbor nodes features to that decision. To this end, we propose a gradient attribution analysis method for the GCN called Node Attribution Method (NAM), which can get the model contribution from not only the central node but also its neighbor nodes to the GCN output. We also propose the Node Importance Visualization (NIV) method to visualize the central node and its neighbor nodes based on the value of the contribution. Moreover, the perturbation analysis method is utilized to verify the efficiency of the NAM based on the citation network datasets. The experimental results indicate that NAM can well learn the contribution of each node to the node classification prediction.

**INDEX TERMS** GCN, Interpret, Attribution Method.

## 1. INTRODUCTION

In recent years, the deep learning model represented by the CNN surpasses human recognition accuracy in the field of the image recognition [1]. However, Different from image where the data are continuous, graph data is discrete. Therefore, the research in the CNN field is difficult to apply to the relational data represented by the graph directly. Although there have been some work converting a graph structure into an image forcibly [2], this method loss the correlation between data, which cannot make good use of graph structure for inference.

In order to solve the problems above, Kipf et al. proposed a GCN [3], which have reached state-of-the-art results in many transductive classification tasks. The convolution operation of each layer in the GCN, in terms of its core idea, essentially performs two-fold information fusions for each node in a given graph. For example, Fig. 1 (a) shows a local graph structure which is composed by a central node $a$ with its one-hop neighbor nodes $\mathcal{N}_{a,1} = \{b, c, d\}$ and its two-hop neighbor node $e$. As the input of the GCN, it including the topology information and node feature information of a given graph. Different letters and colors distinguish different nodes. In this figure, the bar next to each node represents a five-dimensional feature vector. If a dimension value is not zero, the source of that value is labeled by the same color same as the source node, otherwise it is white. In order to obtain the feature $\tilde{a}^{(0)}$ that merged the one-hop neighbor information, the GCN layer performs a weighted summation for the features of $\mathcal{N}_{a,1}$ along each feature dimension. The summation weights are depend on a row vector $\mathcal{L}_a$, which is getting from the Laplacian matrix $\mathcal{L}$ (reflecting the spatial topology of the graph). Fig. 1 (b) illustrates this process. Because this procedure uses the information of $\mathcal{L}$ to perform the weighted summation of the features of $\mathcal{N}_{a,1}$. Thus, we call it as the first-fold fusion of the graph convolutional layer, which fuse the first-order neighbor information of graph structure for each node. Based on the first-fold fusion above, the GCN takes the merged feature $\tilde{a}^{(0)}$ as input and fuses the feature dimension through a fully connected feedforward neural network shared by each node. The purpose of that is to approximate the convolution operation using the high sparsity of the features to achieve the compressive fusion for the feature dimension. Therefore, we call it as the second-fold fusion of the graph convolutional layer, which compress the input feature dimension from high to low for each node. This process is shown in Fig. 1 (c). Note that one GCN layer equals to fuse one-hop neighbors feature information. Therefore, the GCN expand the convolutional receptive field through stacking $K$ GCN layers. Generally K = 2, it means that this GCN model fuse two-hop neighbor features information for each node decision.

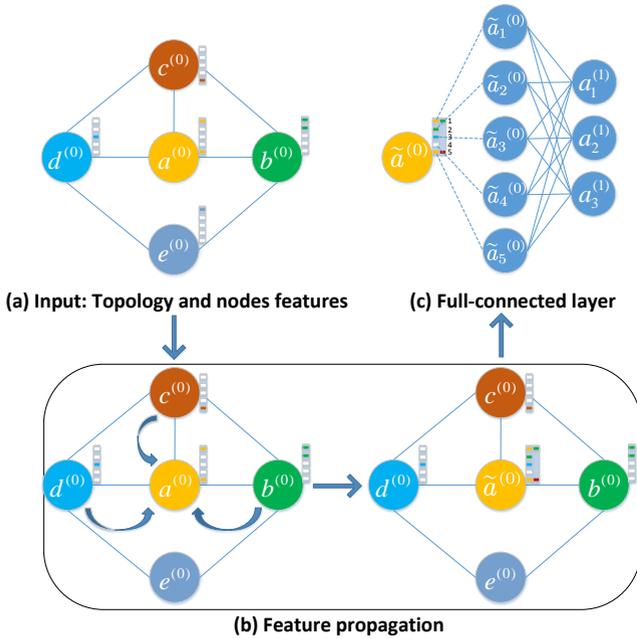

**(a) Input: Topology and nodes features**

**(c) Full-connected layer**

**(b) Feature propagation**

Fig. 1 How does the GCN layer works.

Macroscopically speaking, GCNs are widely used in the field of scene semantic relationship analysis [4]–[8], natural language processing [9], [10], few-shot learning [11], etc. due to its combinatorial generalization characteristics [12]. However, like other deep learning models, GCNs are also have many interpretable problems that need to be explained [13]. It is troublesome to do the attribution analysis for the predictions made by the GCNs. In other words, it is difficult to infer which nodes or feature dimensions best influence the prediction, which is very important in many data analysis problems. Such as the fraud prevention credit prediction system in the financial field. A fraudster uses the adversarial examples of the predictive model to deceive the network to get a high credit score. If the decision-making model cannot be analyzed and interpreted, it is difficult to find and analyze the adversarial attack means in the massive transaction data, so that the enterprise suffers huge economic losses. The difficulty of GCN attribution analysis lies in the fact that GCN realizes the information fusion of graph structure and feature dimension and both of them are the explicit irreversible processes. Therefore, the output obtained by the two information fusion methods is difficult to be disassembled into the form of the information source as shown by the feature vector $\tilde{a}^{(0)}$ in Fig. 1 (b). (Note that the feature information represented by $\tilde{a}^{(0)}$ in different colors is only for explaining the spatial information fusion process of GCN, and it does not mean that the information source is detachable).

As of now, there are few relatively papers on interpretability analysis of the graph network, but there are many network interpretability studies in the fields of DNN and CNN. Such as gradient-based attribution analysis methods, perturbation-based attribution analysis methods and analysis methods of information bottleneck principle etc. The gradient-based attribution analysis methods can calculate the influence of each dimension of the specified input to the decision basing on the network structure quantitatively. For example, using gradient-based attribution methods to analyze CNN [14], the core idea of it is input a data $I_a$(e.g. a picture) into a trained CNN (fixed model parameters), and after $I_a$ passing through the feedforward neural network to get the inactivation output $O_c(I_a)$ of class $c$, then it calculate the gradients $g(I_a, O_c(I_a))$ from the $O_c(I_a)$ to the each dimension of $I_a$ . Especially, $g(I_a, O_c(I_a))$ is the importance coefficient of each input dimension to the predict result. Thus we can obtain the contribution $C_c(I_a, O_c(I_a))$ of each input dimension to the output of CNN by multiply $g_c(I_a, O_c(I_a))$ and $I_a$. To analyze the decision reasons of the CNN intuitively, the heat map visualizations are formed on the value of $C_c(I_a, O_c(I_a))$.

However, the application of these attribution analysis methods to GCN is a non-trivial problem. Firstly, since the convolution of CNN occurs inside an input data, the input data is relatively independent. Therefore, the gradient attribution method can be used to observe the contribution $C_c(I_a, O_c(I_a))$ from the different dimensional features of each input data $I_a$ to the real category output $O_c(I_a)$ independently. In contrast, the convolution of GCN involves the fusion of feature information from the central node $a$ and its K-hop neighbor nodes $\varphi_{a,K}^{(0)}(\varphi \in \mathcal{N}_{a,K})$ , which needs to consider the correlation between the input data. Yet the existing gradient attribution method can only obtain the model contribution $C_c(a^{(0)}, a^{(2)})$ instead of the model contribution $C_c(\varphi_{a,K}^{(0)}, a^{(2)})$ . Secondly, image data has a strong local semantic context information, which uses heat map to highlights the significant influence area and so that it is convenient for us to observe the semantic information of image that which object are most important for the model decision. For example, the real category is a frog, and the area with a larger heat map value is often nearby the frog instead of the background. However, the graph data cannot observe the semantic information through the method of such heat map.

In this work, we come up with a migrating approach to drift the gradient-based attribution analysis method to explain the GCN seamlessly. For this reason, in the aspect of the gradient propagation, we propose a NAM to draw the predictive contribution of each K-hop neighbor nodes, which is working by stratified enumerating all active paths to gain the complete contribution. In the aspect of visualization, we propose a NIV method to observe the contribution of nodes intuitively, thus expert can analyze the decision-making reason of the GCN visually.

Overall, the contributions of our work are:

(1) We extend the existing gradient-based attribution analysis method to explain the GCN and get the model contribution of the central node and its neighbor nodes.

(2) We propose a new visualizable analysis method to demonstrate GCN decision-making reason, which can help expert to visualize the importance degree of the central node and its neighbors to the central node decision making.

(3) We have done many experiments to verify the effectiveness of NAM.

## 2. Preliminaries

In this part, we summarize the development history of GCNs and introduce the basic principles of gradient-based attribution methods briefly, which make a theoretical foundation for the next section.

### 2.1 Graph Convolutional Neural Network

With the flourishing of deep learning, a mass of outstanding works have emerged to the public attention [15]. Among those researches, CNNs [16] are distinguished in the grid-like data such as images. As a more general data structure, graphs can use nodes and edges to encode more information [17], [18], in which the number of neighbor nodes is often not fixed. We can also consider grid-like data as a special case of graph data that each nodes have the same number of neighbor nodes. However, the feature extraction using regular CNNs have to fix neighborhoods numbers and its orders so that the weights of convolutional kernel can shared conveniently. Therefore, it is hard to use regular convolutional operation into graphs directly. In order to promote convolutional-like operation from the Euclidean space to the Non-Euclidean space, the literatures [19], [20] come up with a graph convolutional network, which transforms the Laplacian matrix from time domain to frequency domain for spectral decomposition and parameterize the Eigenvalue to achieve convolutional approximation. However, the above method exists the following disadvantages: (1). Laplacian matrix decomposition takes a lot of time for large images. (2). The neural network forward propagation must calculate $U\Lambda U^T$ for each layer, and the computational complexity is $\mathcal{O}(n^2)$. (3). The convolutional kernel size is the total nodes number of the graph, and there are too many convolutional parameters. Referring to [22], the K-order Chebyshev polynomial is used to approximate the Laplacian matrix, so that the Laplace matrix does not need to be decomposed. Thus, the computational complexity drops to $\mathcal{O}(n)$, and the convolution kernel size is reduced to $m(m \ll n)$, where m is the K-order neighbor of the node. Literature [3] propose a GCN model, which applies a convolutional operation of the Chebyshev first-order approximation to the semi-supervised learning task on undirected graph. In terms of results, it improved the accuracy of classification greatly. Compared to $K$-order Chebyshev polynomial, GCN limits $K$ to one and establishes receptive field of $K$-hop neighbor nodes through stacking the network layer, which further improves the forward propagation efficiency of the neural network and increase the scalability and performance on the large-scale graph classification problem.

The convolutional forward propagation of it is shown in equation (1). Here, $\tilde{A} = A + I_N$ is the sum of the adjacency matrix and the identity matrix. $\tilde{D}$ is the degree matrix of the $\tilde{A}$. $\tilde{D}^{-\frac{1}{2}}\tilde{A}\tilde{D}^{-\frac{1}{2}}$ is an approximate symmetric normalized Laplacian matrix $\mathcal{L}$, the $i$-th row and the $j$-th column of $\mathcal{L}$ indicates the undirected edge weight $e_{ij} = \frac{1}{\sqrt{\deg(i)\deg(j)}}$ between node $i$ and node $j$. $H^{(l)}$ is a processed features matrix for all nodes in a graph that fuse $l$-order neighbors information by the convolutional layers before the layer $l$, $1 \leq l \leq K$. Note that $H^{(0)}$ is the original input data. The symbolic representation mentioned above are all parameters that will be used in first fold informational fusion of a GCN layer. $W^{(l)}$ is the weight of the fully connected part in the $l$-th GCN layer and $b^{(l)}$ is the offset value of this part.

$$H^{(l)} = f\left(\tilde{D}^{-\frac{1}{2}}\tilde{A}\tilde{D}^{-\frac{1}{2}}H^{(l-1)}W^{(l)} + b^{(l)}\right) \qquad (1)$$

### 2.2 Gradient-based Attribution Methods

Given a trained neural network, to analyze the importance of each feature dimension in a specific input in terms of the model output, there are mainly two attribution methods that shown as follows: perturbation-based methods and backpropagation-based (gradient-based) methods. Here we chiefly discuss the gradient-based attribution method.

In order to facilitate the understanding of the gradient attribution method based on backpropagation, we first introduce the structure of the feedforward neural network. equation (2) and equation (3) are the operational relationships between adjacent layers in a feedforward neural network.

$$z_j^{(l+1)} = \sum_{i \in \mathbb{I}_j^{(l+1)}} w_{ij}^{(l,l+1)} x_i^{(l)} + b_j^{(l+1)} \qquad (2)$$

$$x_j^{(l+1)} = \sigma\left(z_j^{(l+1)}\right) \qquad (3)$$

Here, $x_j^{(l)}$ is the activation output of the $j$-th neuron in the $l$-th layer. Note that $l = 0,1,...,L(L$ is the total number of layers in a neural network) and $x^{(0)}$ refers to the input data. Relatively, $z_j^{(l)}$ is the input before activation of the layer $l$ neuron $j$, in which $w_{ij}^{(l-1,l)}$ is the weight of the layer $l-1$ neuron $i$ connected to the layer $l$ neuron $j$, and $b_j^{(l)}$ is the bias of the layer $l$ neuron $j$. $\mathbb{I}_j^{(l)}$ is a set of the neurons in the layer $l-1$ which connect to the layer $l$ neuron $j$. Specifically, in the fully connected layer, $\mathbb{I}_j^{(l)}$ represents all neurons in the layer $l-1$; in the convolutional layer, $\mathbb{I}_j^{(l)}$ represents a set of the neurons in the layer $l-1$, which are connected through the convolution kernel. In addition, $\sigma(\cdot)$ is the activation function, and common ones are ReLU, Tanh, Softmax, Sigmoid etc. In order to simplify the representation, we give an abbreviation formula between adjacent layers for the general neural network that is shown in equation (4). Therefore, if we need to obtain the partial derivative between two adjacent layers in a neural network, we can get the result using equation (5). equation (6) is a more flexible formula of it, which can specify the specific dimension in the formula of derivative.

$$X^{(l+1)} = \sigma(Z^{(l+1)}) = \sigma(W^{l,l+1}X^{(l)}) \quad (4)$$

$$\frac{\partial X^{(l+1)}}{\partial X^{(l)}} = W^{(l,l+1)}\frac{\partial \sigma(Z^{(l+1)})}{\partial Z^{(l+1)}} \quad (5)$$

$$\frac{\partial x_j^{(l+1)}}{\partial x_i^{(l)}} = w_{ij}^{(l,l+1)}\frac{\partial \sigma(z_j^{(l+1)})}{\partial z_j^{(l+1)}} \quad (6)$$

After known the general structure between adjacent neural network layers, we start to introduce the gradient-based attribution method [23]. First, assume that the classifier $\mathfrak{C}$ (such as a neural network) is a linear function so that we can have an elegant interpretation for each input dimensions according to weights. For example, input is a specific picture $I_a$ and output is $O(I_a)$. Where $O(I_a) = W_aI_a + b_a = \sum_i w_{ai}i_a + b_a$, $W_a$ is the weight vector for the input $I_a$. We can regard $w_{ai} \times i_a$ as the real influence of each input dimension to its output. In other words, we can obtain the contribution of each input dimension to model decision. Moreover, if $\mathfrak{C}$ is a highly nonlinear function such as a neural network, a first-order Taylor expansion can be performed at $(I_a, \mathfrak{C}(I_a))$ to obtain a good interpretability like the linear function. If the input $I$ is in the neighborhood of $I_a$, then the expansion is shown in equation (7).

$$O(I_a) \approx \mathfrak{C}(I_a) + \mathfrak{C}'(I_a)(I - I_a)$$
$$= \mathfrak{C}'(I_a)I + \beta_a \quad (7)$$

Where $\mathfrak{C}'(I_a)$ in here is equivalent to $W_a$ in the linear function, which can be calculated by the chain rule of derivatives according to the general structure of the neural network formula mentioned above:

$$\mathfrak{C}'(I_a) = \frac{\partial X^{(L)}}{\partial X^{(0)}}\bigg|_{X=I_a}$$
$$= \prod_{l=0}^{L-1}\frac{\partial X^{(l+1)}}{\partial X^{(l)}}\bigg|_{X=I_a}$$
$$= \prod_{l=0}^{L-1}W^{(l,l+1)}\frac{\partial \sigma(Z^{(l+1)})}{\partial Z^{(l+1)}}\bigg|_{X=I_a} \quad (8)$$

Equation (8) shows the partial derivative structure of general neural networks. When use gradient-based attribution method, we need to utilize the active paths of the neural network as the partial derivative structure, which is the feed forward propagation structure after input a sample. More generally, we can get the partial derivative of any layer output $x_j^{(t)}$ with respect to any layer input $x_i^{(s)}$, where $s < t$. This formula is shown in equation (9).

$$\frac{\partial x_j^{(t)}}{\partial x_i^{(s)}} = \sum_{p\in\mathcal{P}_{st}}\left(\prod_{l\in\mathcal{L}_p}\frac{\partial x_p^{(l+1)}}{\partial x_p^{(l)}}\right)$$
$$= \sum_{p\in\mathcal{P}_{st}}\left(\prod_{l\in\mathcal{L}_p}\left(w_p^{(l,l+1)}\frac{\partial \sigma(z_p^{(l+1)})}{\partial z_p^{(l+1)}}\right)\right) \quad (9)$$

Where $\mathcal{P}_{st}$ is all paths from $x_i^{(s)}$ to $x_j^{(t)}$, $\mathcal{L}_p$ is all layers on path $p$, and the partial derivative values of all paths are accumulated to obtain the influence of $x_i^{(s)}$ on $x_j^{(t)}$.

Note that both the softmax activation and the sigmoid activation may cause the problems of the important attenuation [24]. Therefore, when performing attribution method, those activations are often modified into the linear activation. Meanwhile, the activation of the model last layer is often softmax in the classification task, thus it usually set several inactivation neurons of the last layer as the importance indicator of attribution analysis.

## 3. Interpretation Model

In this section, we first introduce the challenges using gradient-based attribution method to interpret GCN, and then propose two method, NAM and NIV, to migrate existing attribution methods to analyze GCN seamlessly.

### 3.1 Challenges

In this part, starting with the model structure of GCN, we explain the problems will encounter when directly using existing gradient-based attribution methods to interpret the decision-making reason for GCN.

The detailed structure of the forward propagation of the GCN is shown in Fig. 2.

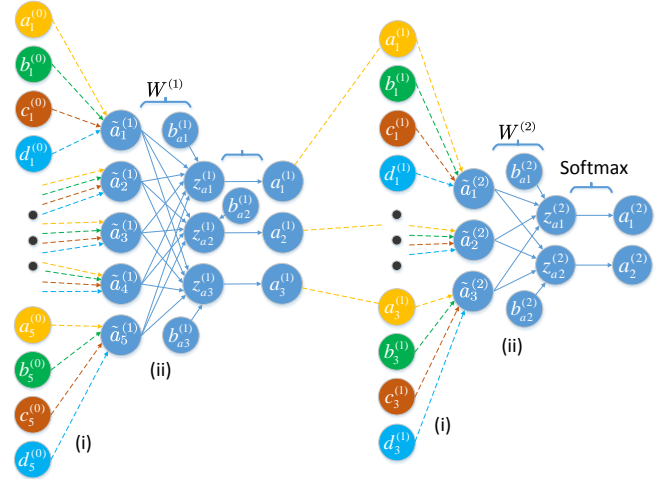

Fig. 2. The detailed structure of the GCN forward propagation. $a$ is the central node for each layer. Left: the first layer of the GCN. Right: the second layer of the GCN. Part (i) represents the first-fold fusion of the GCN in the graph structure. Part (ii) represents the second-fold fusion of the GCN in the feature dimension.

Where $\varphi_i^{(l)}$ indicates the $i$-th feature dimension of node $\varphi$, which is processed after the first $l$ layers ($l = 0$ means it is original input feature) and $\tilde{a}_i^{(l)}$ denotes the $i$-th dimension of fused feature of the central node $a$.

Since existing gradient-based attribution methods does not take into account the correlation between data like the part of Fig. 2 (i), two problems cannot be solved.

Firstly, observing from Fig. 2, consider the feed forward propagation process, to obtain the output of the first layer $a^{(1)}$, which need to know the value of $a^{(0)}$ and $\varphi_{a,1}^{(0)}$. Similarly, to earn the output of the second layer $a^{(2)}$, it also needs to have the information of $a^{(1)}$ and $\varphi_{a,1}^{(1)}$ (i.e. $\varphi_{a,2}^{(0)}$). Therefore, when we calculate the active paths of a model decision-making, different feed forward propagation path of different input sample need to be calculated respectively. For example, if we need to get the model contribution from $a^{(0)}$ to $a^{(2)}$, the existing gradient-based attribution methods can only calculate the partial derivatives that $\frac{\partial a^{(2)}}{\partial a^{(1)}} \times \frac{\partial a^{(1)}}{\partial a^{(0)}}$, while ignore the other active path like $a^{(0)}$ to $\varphi_{a,1}^{(1)}$ to $a^{(2)}$.

Secondly, the existing gradient-based attribution methods aim to analyze the situation that the correspondence relationship is one input sample to one output prediction. Correspondingly, the implementation of the GCN treating different node feature as disparate input sample in a same batch, which can only utilize the existing gradient-based attribution methods to get the model contribution from original input feature $a^{(0)}$ to its model prediction $a_c^{(2)}$ of class $c$, while the model contribution of $\varphi_{a,K}^{(0)}$ to $a_c^{(2)}$ cannot be obtained.

In addition, the existing attribution method usually concentrate on the importance of the feature dimension, while the feature of each node are high dimension and very sparse. Therefore, it is reasonless to make use of feature dimension as the importance unit of the attribution analysis on graph data.

Moreover, the visualization method of the existing attribution method usually utilize the heat map to visualize the result of the attribution contribution. While non-grid data do not have the local semantic context information, thus it is hard to use heat map to visualize graph data directly.

Overall, we separate two types of problems when use the existing gradient-based attribution method: the attribution problems and the visualization problems.

### 3.2 Node Attribution Method

In order to solve the attribution problems mentioned above, we propose NAM to obtain the model contribution of node feature rather than feature dimension, and make it possible to analyze the model contribution of each K-hop neighbors for the central node. NAM also considers multiple active paths to get the relatively complete model contribution value. Those characteristics are guaranteed by three rules that list on below.

#### 3.2.1 Intra-layer calculation rule

For the $l$-th GCN layer $l$, the part (ii) of Fig. 2 is fully connected layer, thus NAM can utilize the existing gradient-based attribution method to calculate the gradient between

$\tilde{a}^{(l)}$ and $a^{(l)}$, which differentiate $a^{(l)}$ with respect to $\tilde{a}^{(l)}$. This rule is shown as equation (10).

$$g_{\tilde{a}}^{(l)} = \frac{\partial a^{(l)}}{\partial \tilde{a}^{(l)}} = W^{(l)} \frac{\partial \sigma(z_a^{(l)})}{\partial z_a^{(l)}} \quad (10)$$

#### 3.2.2 Node attribution rule

Because of the high dimension of node feature and non-semantic of the local context, it is not a wise way to utilize the feature dimension as the importance metric element. We define the node attribution rule to attribute the node contribution. Formally, $\varphi, \varphi'$ are any neighbor nodes within one-hop distance, and $\varphi$ is the central node of this layer. $c$ denotes the class of the ground truth. $g_{\tilde{\varphi}}^{(l)}$ is the gradient between $\tilde{\varphi}^{(l)}$ and $\varphi^{(l)}$, which is introduced in 3.2.1. This rule obtain the summation of the dimension contribution as the layer node contribution. Equation (11) shows this rule.

$$C_c\left(\varphi'^{(l-1)}, g_{\tilde{\varphi}}^{(l)}\right) = \left(\varphi'^{(l-1)}\right)^T \times g_{\tilde{\varphi}}^{(l)} \quad (11)$$

Due to $\varphi'$ can be any neighbor nodes of central node, thus this rule also address the problem that the existing attribution method cannot figure out the node contribution of neighbor nodes in a GCN layer.

#### 3.2.3 Hierarchical calculation rule

To deal with the problem of the incomplete active paths, NAM utilize the intra-layer calculation rule and the node attribution rule to compute the contribution inside a GCN layer, and then connect them by the chain rule of the derivative. Take the Fig. 2 as an example, the model contribution of $a^{(0)}$ is shown as follows.

$$\begin{aligned} C_c\left(a^{(0)}, a^{(2)}\right) &= C_c\left(a^{(0)}, g_{\tilde{a}}^{(1)}\right) \times C_c\left(a^{(1)}, g_{\tilde{a}}^{(2)}\right) \\ &= C_c\left(a^{(0)}, g_{\tilde{a}}^{(1)}\right) \times C_c\left(\varphi^{(1)}, g_{\tilde{a}}^{(2)}\right) \end{aligned} \quad (12)$$

Equation (12) shows the model contribution of $a^{(0)}$ to $a^{(2)}$, which active paths are drawn in Fig. 3.

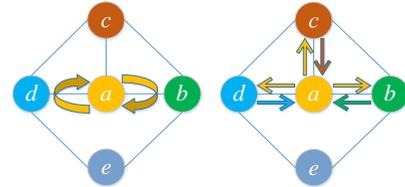

Fig. 3. The active paths from $a^{(0)}$ to $a^{(2)}$. Left: $a^{(0)} \rightarrow a^{(1)} \rightarrow a^{(2)}$. Right: $a^{(0)} \rightarrow \varphi^{(1)} \rightarrow a^{(2)}$.

Through this rule, we can get any model contribution of K-hop neighbor nodes to the central node prediction. The active paths of the model contribution of $\varphi_{a,1}^{(0)}$ is illustrated in Fig. 4.

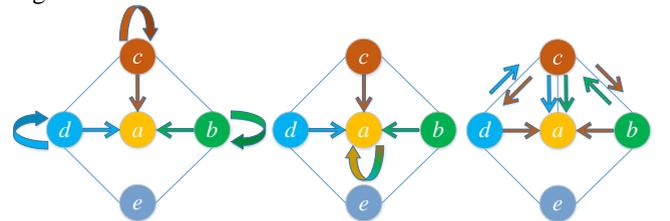

Fig. 4. The active paths from $\varphi_{a,1}^{(0)}$ to $a^{(2)}$. Left: $\varphi_{a,1}^{(0)} \rightarrow \varphi_{a,1}^{(1)} \rightarrow a^{(2)}$; Middle: $\varphi_{a,1}^{(0)} \rightarrow a^{(1)} \rightarrow a^{(2)}$; Right: $\varphi_{a,1}^{(0)} \rightarrow$

$$\varphi'^{(1)}_{a,1} \rightarrow a^{(2)}.$$

Where $\varphi_{a,1}$, $\varphi'_{a,1}$ are two different neighbor nodes of central node $a$. The formula of it is shown in equation (13).

$$C_c\left(\varphi^{(0)}_{a,1}, a^{(2)}\right) = C_c\left(\varphi^{(0)}_{a,1}, g^{(1)}_{\bar{\varphi}_{a,1}}\right) \times C_c\left(\varphi^{(1)}_{a,1}, g^{(2)}_{\bar{a}}\right)$$
$$+ C_c\left(\varphi^{(0)}_{a,1}, g^{(1)}_{\bar{a}}\right) \times C_c\left(a^{(1)}, g^{(2)}_{\bar{a}}\right)$$
$$+ C_c\left(\varphi^{(0)}_{a,1}, g^{(1)}_{\bar{\varphi}_{a,1}}\right) \times C_c\left(\varphi^{(1)}_{a,1}, g^{(2)}_{\bar{a}}\right) \quad (13)$$

Using the same method, the active paths of the model contribution of $\varphi^{(0)}_{a,2}$ is illustrated in Fig. 5.

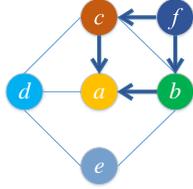

Fig. 5. The active paths from $\varphi^{(0)}_{a,2}$ to $a^{(2)}$, which is $\varphi^{(0)}_{a,2} \rightarrow \varphi^{(1)}_{a,1} \rightarrow a^{(2)}$.

Equation (14) shows the contribution of $\varphi^{(0)}_{a,2}$.

$$C_c\left(\varphi^{(0)}_{a,1}, a^{(2)}\right) = C_c\left(\varphi^{(0)}_{a,2}, g^{(1)}_{\bar{\varphi}_{a,1}}\right) \times C_c\left(\varphi^{(1)}_{a,1}, g^{(2)}_{\bar{a}}\right) \quad (14)$$

# 4. Experimental Studies

In this section, we validate the validity of NAM on the node classification task of the transductive learning same as GCN settings. Experimental results show that NAM is more efficient than other attribution method on the GCN interpretation. Moreover, we visualize the contribution of each $K$-hop neighbor nodes via NIV, which provide an intuitive way to analyze the decision-making reason of the GCN.

### 4.1 Evaluate the validity of NAM

In order to validate the validity of NAM, we performed a perturbation experiment on three citation datasets: those are the Cora, Citeseer and Pubmed[25] that details are summarized in Table 1.

Table 1: Summary of the datasets used in our experiments[26].

| Dataset | Cora | Citeseer | Pubmed |
|---|---|---|---|
| #Nodes | 2708 | 3327 | 19717 |
| #Features | 1433 | 3703 | 500 |
| #Classes | 7 | 6 | 3 |
| #Edges | 5429 | 4732 | 44338 |
| #Training Nodes | 140 | 120 | 60 |
| #Validation Nodes | 500 | 500 | 500 |
| #Test Nodes | 1000 | 1000 | 1000 |
| #Label Rate | 0.052 | 0.036 | 0.003 |
| Average Node Degree | 4 | 5 | 6 |

The verification method is to observe the accuracy of model prediction results by deleting $|D_v|$ neighbor nodes of each central node $v$ according to an increasing percentage $p$ multiplied by the number of the $K$-hop neighbor nodes $|\mathcal{N}_{v,K}|$ i.e. $|D_v| = |\mathcal{N}_{v,K}| \times p$. Here, along with the percentage $p$ from large to small, we compare the decreasing speed of accuracy between NAM and random method (RM) through deleting $|D_v|$ neighbor nodes. The experimental results on the Cora dataset are shown in Fig. 6.

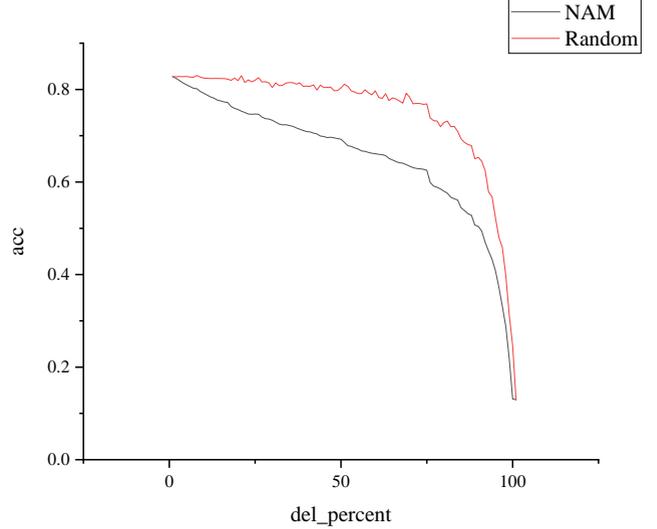

Fig. 6. The predictive accuracy curves with the decreasing neighbor nodes.

It can be seen from Fig. 6 that compared with the RM, the rate of the NAM accuracy curve decreases faster than random significantly. That is to say, in the case of deleting $|D_v|$ neighbor nodes, the NAM is more efficient than the RM deletion node. This validate the effectiveness of NAM in analyzing the influence of neighbor nodes on the decision-making of GCN.

### 4.2 NIV

Just as the CNN uses the heat map to analyze the attribution results, we visualize the two-hop neighbor nodes of the central node $v$ to analyze the decision-making reason of the GCN intuitively. Here, the node size encodes the contribution magnitude and the color represents the result of the predictive classification category. Fig. 7 is the visualization analysis result of one-hop neighbor nodes predicting on the node 1358 of the Cora dataset. Fig. 8 visualize the contribution results on the node 1422 of the Citeseer dataset. Fig. 9 shows the contribution results on the node 11450 of the Pubmed dataset. It can be seen that the most influential node in Fig. 7 is the node 919, the most influential in Fig. 8 is the node 2064, and the most influential in Fig. 9 is the node 14751.

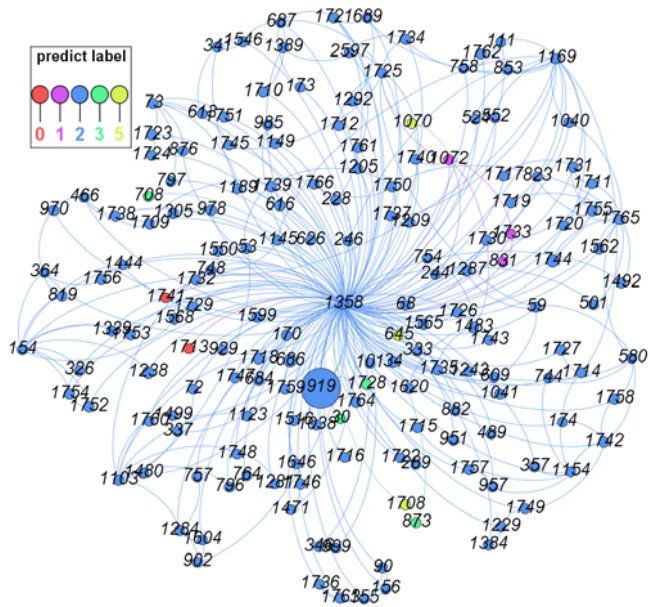

Fig. 7. Visualizing the contributions of one-hop neighbor nodes of node 1358 in Cora dataset.

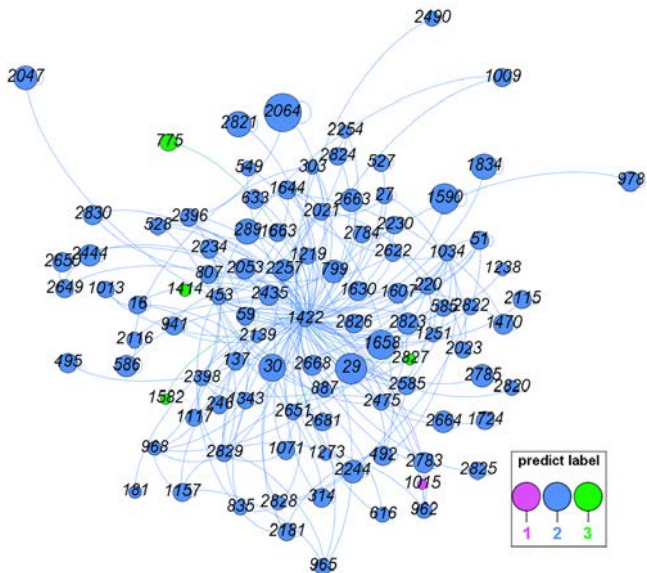

Fig. 8. Visualizing the contributions of one-hop neighbor nodes of node 1422 in Citeseer dataset.

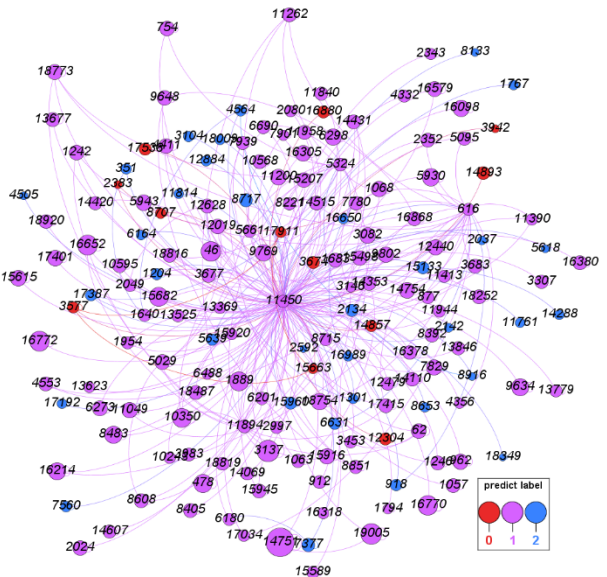

Fig. 9. Visualizing the contributions of one-hop neighbor nodes of node 11450 in Pubmed dataset.

## 5. Conclusions

In this work, we propose a method to extend the existing gradient-based attribution method to analyze the decision-making reason of the GCN seamlessly. The NAM is used to obtain more comprehensive contributions for each node, and the NIV is used to visualize the node contributions to observe the importance of the node to the decision. In addition, the effectiveness of the NAM method is validated in experiment.

## Acknowledgements